\documentclass[11pt,a4paper]{article}
\usepackage[hyperref]{acl2017}
\usepackage{times}
\usepackage{latexsym}
\usepackage{graphicx}
\usepackage{url}
\usepackage{enumitem} 
\usepackage[normalem]{ulem}  
\usepackage{comment}

\newcommand{\ed}[1]{{\color{black} #1}}

\aclfinalcopy 

  \title{Emotion Intensities in Tweets}

\author{\hspace*{-4mm} Saif M. Mohammad\\
\hspace*{-4mm} Information and Communications Technologies\\
	    \hspace*{-4mm} National Research Council Canada\\
        \hspace*{-4mm} Ottawa, Canada\\
	    \hspace*{-4mm} {\tt saif.mohammad@nrc-cnrc.gc.ca} \\\And
  \hspace*{13mm} Felipe Bravo-Marquez \\
  \hspace*{13mm} Department of Computer Science\\ 
  \hspace*{13mm} The University of Waikato\\
\hspace*{13mm} Hamilton, New Zealand\\
  \hspace*{13mm} {\tt fbravoma@waikato.ac.nz} \\}

\date{}

\begin{document}
\maketitle
\begin{abstract}
This paper examines the task of detecting intensity of emotion from text.
\ed{We create the first datasets of tweets annotated for anger, fear, joy, and sadness intensities.} We use a technique called best--worst scaling (BWS)
that improves annotation consistency and obtains reliable fine-grained scores.
We show that emotion-word hashtags often
impact emotion intensity, usually conveying a more intense emotion.
Finally, we create a benchmark regression system and conduct experiments to determine: which features are useful for detecting emotion intensity;
and, the extent to which two emotions are similar in terms of how they manifest in language.
\end{abstract}

\section{Introduction}
\setitemize[0]{leftmargin=*}
\setenumerate[0]{leftmargin=*}
We use language to communicate not only the emotion we are feeling but also the intensity of the emotion. For example, our utterances can convey that we are very angry, slightly sad, absolutely elated, etc. Here, {\it intensity} refers to the degree or amount of an emotion such as anger or sadness.\footnote{Intensity is different from {\it arousal}, which refers to the extent to which an emotion is calming or exciting.} 
Natural language applications can benefit from knowing both the class of emotion and its intensity.
For example, a commercial customer satisfaction system would prefer to focus first on instances of significant frustration or anger, as opposed to instances of minor inconvenience. However, 
most work on automatic emotion detection 
has focused on categorical
classification (presence of anger, joy, sadness, etc.).
A notable obstacle in developing automatic affect intensity systems is the lack of suitable annotated data. 
Existing affect datasets are predominantly categorical.
Annotating instances for degrees of affect is a substantially more difficult undertaking: 
respondents are presented with  greater cognitive load and it is particularly hard to ensure consistency
(both across responses by different annotators and within the responses produced by an individual annotator). 

{\it Best--Worst Scaling (BWS)} 
is an annotation scheme that  addresses these limitations
\cite{Louviere_1991,Louviere2015,maxdiff-naacl2016,KiritchenkoM2017bwsvsrs}.
Annotators are given $n$ items (an $n$-tuple, where $n > 1$ and commonly $n= 4$). They are asked which item is the {\it best} (highest in terms of the property of interest) and which is the {\it worst} (lowest in terms of the property of interest).
When working on $4$-tuples, best--worst annotations are particularly efficient because each best and worst annotation will reveal the order of
five of the six item pairs. For example, for a 4-tuple with items A, B, C, and D, if A is the best, and D is the worst, then A $>$ B, A $>$ C, A $>$ D, B $>$ D, and C $>$ D.

BWS annotations for a set of $4$-tuples can be easily converted into real-valued scores of association 
between the items and the property of interest \cite{Orme_2009,flynn2014}. 
It has been empirically shown that annotations for $2N$ $4$-tuples is sufficient for obtaining reliable scores (where N is the number of items) \cite{Louviere_1991,maxdiff-naacl2016}.\footnote{At its limit, when $n=2$, BWS becomes a {\it paired comparison} \cite{thurstone1927law,david1963method}, but then a much larger set of tuples need to be annotated (closer to $N^2$).}
The little work using BWS in computational linguistics has focused on words \cite{jurgens-EtAl:2012:STARSEM-SEMEVAL,maxdiff-naacl2016}. It is unclear whether the approach can be scaled up to larger textual units such as sentences.


Twitter has a large and diverse user base, which
entails rich 
textual content, including non-standard language such as emoticons, emojis, creatively spelled words ({\it happee}), and hashtagged words ({\it \#luvumom}). 
Tweets are often used to convey one's emotions, opinions towards products, and stance over
issues. Thus, automatically detecting emotion intensities in tweets has many applications, including: tracking
brand and product perception, tracking support for issues and policies, tracking public health and well-being,
and disaster/crisis management.

In this paper, we present work on detecting intensities (or degrees) of emotion in tweets. 
Specifically, given a tweet and an emotion X, the goal is to determine the intensity or degree of emotion X felt by the speaker---a real-valued score between 0 and 1.{\footnote{\ed{Identifying intensity of emotion evoked in the reader, or intensity of emotion felt by an entity mentioned in the tweet, are also useful, and left for future work.}}
A score of 1 means that the speaker feels the highest amount of emotion X.
A score of 0 means that the speaker feels the lowest amount of emotion X.
We annotate a dataset of tweets for intensity of emotion using best--worst scaling and crowdsourcing.
The main contributions of this work are summarized below:\\[-20pt]
\begin{itemize}
\item We formulate and develop the task of detecting emotion intensities in tweets.\\[-20pt]
\item We create four datasets of tweets annotated for intensity of anger, joy, sadness, and fear, respectively. These are the first of their kind.\footnote{\ed{We have also begun work on creating similar datasets annotated for other emotion categories. We are also creating a dataset annotated for valence, arousal, and dominance. 
}
}\\[-20pt] 
\item We show that Best--Worst Scaling can be successfully applied for
annotating sentences (and not just words). We hope that this will encourage the use of BWS more widely, producing more reliable natural language annotations.\\[-20pt]
\item 
We annotate both tweets and a hashtag-removed version of the tweets. 
We analyse the impact of hashtags on emotion intensity.\\[-20pt]
\item We create a regression system, \ed{{\it AffectiveTweets Package}}, to automatically determine emotion intensity.\footnote{\ed{https://github.com/felipebravom/AffectiveTweets}} We show the extent to which 
 various features help determine emotion intensity. 
The system 
is released as an open-source package for the Weka workbench.\\[-20pt] 
\item We conduct experiments to show the extent to which two emotions are similar as per their manifestation in language, by showing
how predictive the features for one emotion are of another emotion's intensity.\\[-16pt]
\item We provide data for a new shared task \ed{WASSA-2017 Shared Task on  Emotion Intensity.\footnote{\label{STW}\ed{http://saifmohammad.com/WebPages/EmotionIntensity-SharedTask.html}} The competition is organized on a CodaLab website, where participants can upload their submissions, and the leaderboard reports the results.\footnote{https://competitions.codalab.org/competitions/16380}
Twenty-two teams participated. 
A description of the task,  details of participating systems, and results are available in  \citet{MohammadB17wassa}.}\footnote{\ed{Even though the 2017 WASSA shared task has concluded, the CodaLab competition website is kept open.
 Thus the best results obtained by any system on the 2017 test set can be found on the CodaLab leaderboard.}}\\[-20pt]
\end{itemize}
\noindent 
All of the data, annotation questionnaires, evaluation scripts, regression code, and interactive visualizations of the data are made freely available \ed{on the shared task website}.\footnotemark[\getrefnumber{STW}]


\section{Related Work}

Psychologists have argued that  some emotions are more basic than others  \cite{Ekman92,Plutchik80,Parrot01,frijda1988laws}.  However, they disagree on which emotions (and how many) should be classified as basic emotions---some propose 6, some 8, some 20, and so on.
Thus, most efforts in automatic emotion detection have focused 
on a handful of emotions, especially since
manually annotating text for a large number of emotions is arduous. 
\ed{Apart from these categorical models of emotions, certain dimensional models of emotion have also been proposed. The most popular among them, Russell's circumplex model, asserts that all emotions are made up of two core dimensions: valence and arousal \cite{russell2003core}.} 
\ed{In this paper, we describe work} on four emotions that are the most common amongst the many proposals for basic emotions: anger, fear, joy, and sadness. \ed{However, we 
have also begun work on other affect categories, as well as on valence and arousal.}

The vast majority of emotion annotation work provides discrete binary labels to the text instances (joy--nojoy, fear--nofear, and so on) \cite{AlmRS05,AmanS07,brooks2013statistical,NeviarouskayaPI09,Bollen2009}.
The only annotation effort that provided scores for degree of emotion is  by \citet{SemEval2007} as part of one of the SemEval-2007 shared task. 
Annotators were given newspaper headlines and asked to provide scores between 0 and 100 via slide bars in a web interface. 
It is difficult for humans to provide direct scores at such fine granularity.   A common problem is inconsistency in annotations. 
One annotator might assign a score of 79 to a piece of text, whereas another annotator may assign a score of 62 to the same text. It is also common that the same annotator assigns different scores to the same text instance 
at different points in time.
Further, annotators often have a bias towards different parts of the scale, known as {\it scale region bias}.

{\it Best--Worst Scaling (BWS)} 
was developed by \citet{Louviere_1991}, building on some ground-breaking research in the 1960’s in mathematical psychology and psychophysics by Anthony A. J. Marley and Duncan Luce.
\ed{\citet{KiritchenkoM2017bwsvsrs} show through empirical experiments that BWS produces more reliable fine-grained scores than scores obtained using rating scales.}
Within the NLP community, Best--Worst Scaling (BWS) has thus far been used only to annotate words: for example, for creating datasets for relational similarity \cite{jurgens-EtAl:2012:STARSEM-SEMEVAL}, word-sense disambiguation \cite{Jurgens2013EmbracingAA}, word--sentiment intensity \cite{Kiritchenko2014},
and phrase sentiment composition \cite{maxdiff-naacl2016}. 
However, in this work we use BWS to annotate whole tweets for degree of emotion. With BWS we address the challenges of direct scoring, and produce more reliable emotion intensity scores.  Further,  this will be the first dataset with emotion 
scores for {\it tweets}.

Automatic emotion classification has been proposed for many different kinds of texts,
including tweets \cite{summa2016microblog,Mohammad12,Bollen2009,AmanS07,brooks2013statistical}.
However, there is little work on emotion regression other than the three submissions to the 2007 SemEval task \cite{SemEval2007}.

\section{Data}

For each of the four focus emotions, our goal was to create a dataset of tweets such that:\\[-18pt]
\begin{itemize}
\item The tweets are associated with various intensities (or degrees) of emotion.\\[-20pt]
\item Some tweets have words clearly indicative of the focus emotion and some tweets do not.\\[-18pt]
\end{itemize}
A random collection of tweets is likely to have a large proportion of tweets not associated with the focus emotion, and thus annotating all of them for intensity of emotion is sub-optimal. To create a dataset of tweets rich in a particular emotion, we use the following methodology.



For each emotion X, we select 50 to 100 terms that are associated with that emotion at different intensity levels. For example, for the anger dataset, we use the terms: {\it angry, mad, frustrated, annoyed, peeved, irritated, miffed, fury, antagonism,} and so on. For the sadness dataset, we use the terms: {\it sad, devastated, sullen, down, crying, dejected, heartbroken, grief, weeping,} and so on. We will refer to these terms as the {\it query terms}. 

We identified the query words for an emotion by first searching 
the {\it Roget's Thesaurus} to find categories that had the focus emotion word (or a close synonym) as the head word.\footnote{The {\it Roget's Thesaurus}
groups words into about 1000 categories.
The head word is the word 
that best represents the meaning of the words within the category.
The categories chosen  were:
 900 Resentment (for anger),  860 Fear (for fear), 836 Cheerfulness (for joy), and 837 Dejection (for sadness).}
We chose all 
words listed within these categories to be the query terms for the corresponding focus emotion.
We polled the Twitter API for tweets that included the query terms.
We discarded retweets (tweets that start with RT) and
tweets with urls.
We created a subset of the remaining tweets 
by:\\[-18pt]
\begin{itemize}
\item selecting at most 50 tweets per query term.\\[-20pt]
\item selecting at most 1 tweet for every tweeter--query term combination.\\[-18pt]
\end{itemize}
\noindent Thus, the {\it master set of tweets} is 
\ed{not heavily skewed towards some tweeters or query terms.} 

To study the impact of emotion word hashtags on the intensity of the whole tweet, 
\ed{we identified} tweets that had a query term in hashtag form towards the end of the tweet---\ed{specifically, within the trailing portion of the tweet made up solely of hashtagged words.}
\ed{We created copies of these tweets and then 
removed
the hashtag query terms 
from the copies. The updated tweets were then added to the master set.
Finally, our master set \ed{of 7,097 tweets} includes:}\\[-20pt]
\begin{enumerate}
\item {\it Hashtag Query Term Tweets (HQT Tweets)}:\\ \ed{1030 tweets with a query term in the form of a hashtag (\#$<$query term$>$) in the trailing portion of the tweet};\\[-20pt]
\item {\it No Query Term Tweets (NQT Tweets)}:\\ \ed{1030 tweets 
that are copies of `1', but with the hashtagged query term removed;}\\[-18pt]
\item  {\it Query Term Tweets (QT Tweets)}:\\ \ed{5037 tweets that include:\\ a. tweets that contain a query term in the form of a word (no \#$<$query term$>$) \\ b. tweets with a query term in hashtag form followed by at least one non-hashtag word.} 
\end{enumerate}
The master set of tweets was then manually annotated 
for intensity of emotion. Table \ref{tab:tdt} shows a breakdown by emotion.

\subsection{Annotating with Best--Worst Scaling}

We followed the procedure described in \citet{maxdiff-naacl2016} to obtain BWS annotations. 
For each emotion, the annotators were presented with four tweets at a time (4-tuples) and asked to select the speakers of the tweets with the highest and lowest emotion intensity.
$2\times N$ (where $N$ is the number of tweets in the emotion set) distinct 4-tuples were randomly generated in such a manner that
each item is seen in eight different 4-tuples, 
and no pair of items occurs in more than one 4-tuple.
\ed{We will refer to this as {\it random maximum-diversity selection (RMDS)}. RMDS maximizes the number of unique items that  each  item  co-occurs with  in the 4-tuples. After BWS annotations, this in turn leads to direct comparative ranking information for the maximum number of pairs of items.\footnote{\ed{In combinatorial mathematics, {\it balanced incomplete block design} refers to creating blocks  (or tuples) of a handful items from a set of $N$ items such that each item occurs in the same number of blocks (say $x$) and each pair of distinct items occurs in the same number of blocks (say $y$), where $x$ and $y$ are integers $ge$ 1 \cite{yates1936incomplete}. The set of tuples we create have similar properties, except that since we create only $2N$ tuples, pairs of distinct items either never occur together in a 4-tuple or they occur in exactly one 4-tuple.}}

It is desirable for an item to occur in sets of 4-tuples such that the maximum intensities in those 4-tuples are spread across the range from low intensity to high intensity, as then the proportion of times an item is chosen as the best is indicative of its intensity score. Similarly, it is desirable for an item to occur in sets of 4-tuples such that the minimum intensities are spread from low to high intensity. However, since the intensities of items are not known beforehand, RMDS is used.}

Every 4-tuple was annotated by three independent annotators.\footnote{\citet{maxdiff-naacl2016} showed that using just three annotations per 4-tuple produces highly reliable results. Note that since each tweet is seen in eight different 4-tuples, we obtain $8 \times 3 = 24$ judgments over each tweet.} 
The questionnaires used 
were developed 
through internal discussions and pilot annotations. A sample questionnaire 
is shown in the Appendix (A.1).

The 4-tuples of tweets were uploaded on the crowdsourcing platform, CrowdFlower.
About 5\% of the data was annotated internally beforehand (by the authors). These questions are referred to as gold questions. 
The gold questions are interspersed with other questions.
If one gets a gold question wrong, they are immediately notified of it. If one's accuracy on the gold questions falls below 70\%, they are refused further annotation, and all of their annotations are discarded. This serves as a mechanism to avoid malicious  annotations.\footnote{\ed{In case more than one item can be reasonably chosen as the best (or worst) item, then more than one acceptable gold answers are provided. The goal with the gold annotations is to identify clearly poor or malicious annotators. In case where two items are close in intensity, we want the crowd of annotators to indicate, through their BWS annotations, the relative ranking of the items.}}

The BWS responses were translated into 
scores 
by a simple calculation \cite{Orme_2009,flynn2014}: For
each item, the score is the percentage of times the item was chosen as having the most intensity minus the percentage of times the item was chosen as having the least intensity.\footnote{\ed{\citet{maxdiff-naacl2016} provide code for generating tuples from items using RMDS, as well as code for generating scores from BWS annotations: http://saifmohammad.com/WebPages/BestWorst.html}}
The scores range from $-1$ 
to $1$. 
Since degree of emotion is a unipolar scale, we linearly transform the the $-1$ to $1$ scores to scores in the range 0 to 1. 

\begin{table}[t!]
\begin{center}
\small{
\begin{tabular}{lrrrr}
\hline {\bf Emotion} & {\bf Train} & {\bf Dev.} &{\bf Test} &{\bf All}\\ \hline
anger &857 &84 &760 &1701\\
fear &1147 &110 &995 &2252\\
joy  &823 &74 &714 &1611\\
sadness &786 &74  &673 &1533\\ \hline
{\bf All}  &3613 &342 &3142 &7097\\
\hline
\end{tabular}
}
\caption{\label{tab:tdt} {The number of instances in the Tweet Emotion Intensity dataset.}
}
\vspace*{-4mm}
\end{center}
\end{table}

\subsection{Training, Development, and Test Sets}

We refer to the newly created emotion-intensity labeled data as the {\it Tweet Emotion Intensity Dataset}.
The dataset is partitioned
into training, development, and test sets for machine learning experiments (see Table \ref{tab:tdt}). 
For each emotion, we chose to include about 50\% of the tweets in the training set, about 5\% in the development set, and about 45\% in the test set. Further, we made sure that an NQT tweet is in the same partition as the HQT tweet it was created from.
See Appendix (A.4) for  details of an interactive visualization of the data.



\section{Reliability of Annotations}
One cannot use standard inter-annotator agreement measures to determine quality of BWS annotations because the disagreement that arises when a tuple has two items that are  close in emotion intensity is a useful signal for BWS. For a given 4-tuple, if respondents are not able to consistently identify the tweet that has  highest (or lowest) emotion intensity, then the disagreement will lead to the two tweets obtaining scores that are close to each other, which is the desired outcome. Thus a different measure of quality of annotations must be utilized.

A useful measure of quality is reproducibility of the end result---if repeated independent manual annotations from multiple respondents result in similar intensity rankings (and scores), then one can be confident that the scores capture the true emotion intensities. 
To assess this reproducibility, we calculate average {\it split-half reliability (SHR)}, \ed{a commonly used approach to determine consistency
\ed{\cite{kuder1937theory,cronbach1946case}}. 
The intuition behind SHR is as follows.}   
All 
annotations for an item (in our case, tuples) are randomly split into two halves. Two sets of scores are produced independently from the two halves. 
Then the correlation between the two sets of scores is calculated. If the annotations are of good quality, then the correlation between the two halves will be high.

\ed{Since each tuple in this dataset was annotated by three annotators (odd number), we  calculate SHR by randomly placing one or two annotations per tuple in one bin and the remaining (two or one) annotations for the tuple in another bin. Then two sets of intensity scores (and rankings) are calculated  from the  annotations in each of the two bins. The process is repeated 100 times and the correlations across the  two sets of rankings and intensity scores are averaged.
Table \ref{tab:shr} shows the split-half reliabilities for the anger, fear, joy, and sadness tweets in the Tweet Emotion Intensity Dataset.\footnote{\ed{Past work has found the SHR for sentiment intensity annotations for words, with 8 annotations per tuple, to be 0.98 \cite{Kiritchenko2014}. In contrast, here SHR is calculated from 3 annotations, for emotions, and from whole sentences. SHR determined from a smaller number of annotations and on more complex annotation tasks are expected to be lower.}}  Observe that for fear, joy, and sadness datasets, both the Pearson correlations and the Spearman rank correlations lie between 0.84 and 0.88, indicating a high degree of reproducibility. However, the correlations are slightly lower for anger indicating that it is relative more difficult to ascertain the degrees of anger of speakers from their tweets.
Note that SHR indicates the quality of annotations obtained when using only half the number of annotations. The correlations obtained when repeating the experiment  with three annotations for each 4-tuple is expected to be even higher. 
Thus the numbers shown in Table \ref{tab:shr} are a lower bound on the quality of annotations obtained with three annotations per 4-tuple.
}

\begin{table}[t!]
\begin{center}
{\small
\begin{tabular}{lrrlrrlrrlr}
\hline
Emotion		&Spearman	&Pearson\\\hline
anger		&0.779		&0.797\\
fear		&0.845		&0.850\\
joy			&0.881		&0.882\\
sadness		&0.847		&0.847\\
\hline
\end{tabular}
\caption{\label{tab:shr} {\ed{Split-half reliabilities (as measured by Pearson correlation and Spearman rank correlation) for the anger, fear, joy, and sadness tweets in the Tweet Emotion Intensity Dataset.}}}
}
\end{center}
\vspace*{-3mm}
\end{table}

\section{Impact of Emotion Word Hashtags on Emotion Intensity}
Some studies have shown that emoticons tend to be redundant in terms of the sentiment 
\cite{go2009twitter,MohammadSemEval2013}. 
That is, if we remove a smiley face, `:)', from a tweet, we find that the rest of the tweet still conveys a positive sentiment. 
Similarly, it has been shown that hashtag emotion words are also somewhat redundant in terms of the class of emotion being 
conveyed by the rest of the tweet \cite{Mohammad12}. For example, removal of `\#angry' from the tweet below leaves a tweet that still conveys anger.\\[-15pt]
\begin{quote}
\it This mindless support of a demagogue needs to stop. \#racism \#grrr \#angry\\[-15pt]
\end{quote}
\noindent However, it is unclear what impact such emotion word hashtags have on the {\it intensity} of emotion. 
In fact, there exists no prior work to systematically study this. One of the goals of creating this 
dataset and including HQT--NQT tweet pairs, is to allow for exactly such an investigation.\footnote{See Appendix (A.2) for further discussion on how emotion word hashtags have been used in prior research.}

\begin{figure}[t]
\centering
 \includegraphics[width=2.9in]{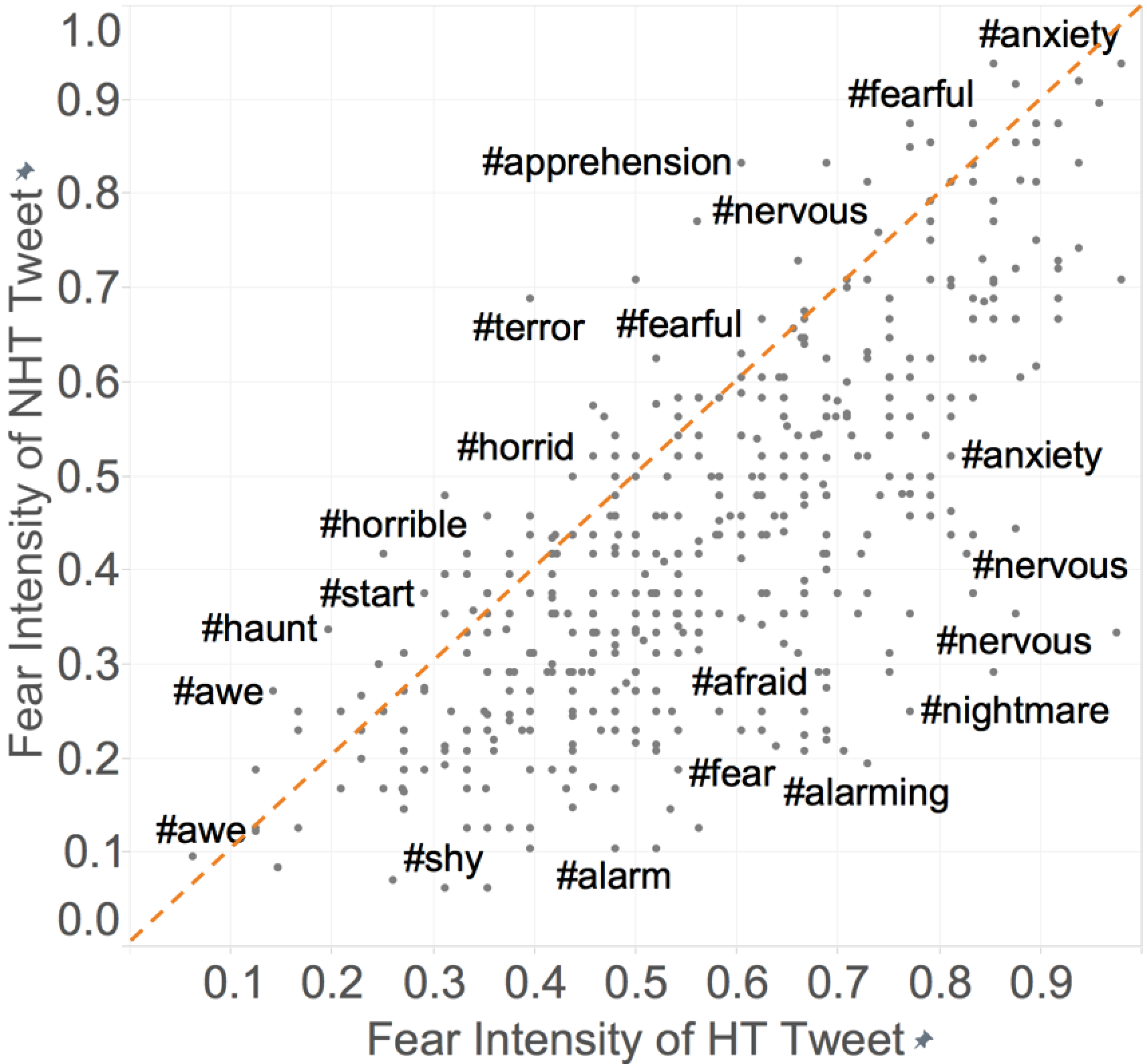}
\caption{The scatter plot of fear intensity of HQT tweet vs.\@ corresponding NQT tweet. \ed{As per space availability, some points are labeled with the relevant hashtag.}}
\label{fig:fear-scatter}
\end{figure}

\begin{table*}[t!]
\begin{center}
\small{
\begin{tabular}{lrrrrrrrrrr}
\hline &\bf No.\@ of HQT--NQT &  & \multicolumn{3}{c}{\bf \% Tweets Pairs}  & & \multicolumn{4}{c}{\bf Average Emotion Intensity Score}\\ 
{\bf Emotion} &\bf Tweet Pairs & &Drop &Rise &None & &HQT tweets& NQT  tweets
&{Drop}	&{Rise}\\ \hline
anger 		&282 	& &76.6 &19.9 &3.4 & &0.58 &0.48 &0.15 &0.07\\
fear 		&454	& &86.1 &13.9 &4.4 & &0.57 &0.43 &0.18 &0.07\\
joy  		&204	& &71.6 &26.5 &1.9 & &0.59 &0.50 &0.15 &0.09\\
sadness 	&90		& &85.6 &11,1 &3.3 & &0.65 &0.49 &0.19 &0.05\\ \hline
{\bf All}  	&1030	& &78.6 &17.8 &3.6 & &0.58 &0.47 &0.17 &0.08\\
\hline
\end{tabular}
}
\caption{\label{tab:pairsimpact} {The impact of removal of emotion word hashtags on the emotion intensities of tweets.}
}
\end{center}
\end{table*}

We analyzed the scores in our dataset 
to create scatter plots where each point corresponds to a HQT--NQT tweet pair, the x-axis is the emotion intensity score of the HQT tweet, and the y-axis is the score of the NQT tweet.
Figure \ref{fig:fear-scatter} shows the scatter plot for the fear data. 
We observe that in a majority of the cases, the points are on the lower-right side of the diagonal, indicating that the removal of the emotion word hashtag causes the emotion intensity of the tweet to drop. 
However, we do see a number of points on the upper-left side of the diagonal (indicating a rise), and some exactly on the diagonal (indicating no impact), due to the removal of a hashtag.
Also observe that the removal of a hashtag can result in a drop in emotion scores for some tweets, but a rise for others (e.g., see the \ed{three labeled} points for {\it \#nervous} in the plot).
We observe a similar pattern for other emotions as well (plots not shown here).
Table \ref{tab:pairsimpact} summarizes these results by showing the percentage of times the three outcomes occur for each of the emotions.

The table also shows that the average scores of HQT tweets 
and NQT tweets. The difference between 0.58 and 0.47 is statistically significant.\footnote{Wilcoxon signed-rank test at 0.05 significance level.}  The last two columns show that when there is a drop in score on removal of the hashtag, the average drop  is  about 0.17 (17\% of the total range 0--1), whereas when there  is a rise, the average rise is  0.08 (8\% of the total range).
These results show that  emotion word hashtags are often {\it not} redundant  with the rest of tweet in terms of what they bring to bear at the overall emotion intensity. Further, even though it is common for many of these hashtags to increase the emotion intensity, there is a  more complex interplay between the text of the tweet  and the hashtag which determines the directionality and magnitude of the impact on emotion intensity. 
For instance, we often found that 
if the rest of the tweet  clearly indicated the presence of an emotion \ed{(through another emotion word hashtag, emojis, or through the non-hashtagged words)}, then the emotion word hashtag had only a small impact on the  score.\footnote{Unless the hashtag word itself is associated with very low emotion intensity (e.g., \#peeved with anger), in which case, there was a drop in perceived emotion intensity.}


However, if the rest of the tweet is under-specified in terms of the emotion of the speaker, then the emotion word hashtag markedly increased the perceived emotion intensity. 
We also observed  patterns unique to particular emotions. For example, when  judging degree of fear of a speaker, lower scores were assigned when the speaker used a hashtag that indicated some outward judgment.\\[-8pt] 

\noindent \hspace*{6mm} {\it @RocksNRopes Can't believe how rude\\
\hspace*{6mm} your cashier was.}
\hspace*{20mm} fear: 0.48\\[-5pt]

\noindent \hspace*{6mm} {\it @RocksNRopes Can't believe how rude\\ 
\noindent \hspace*{6mm} your cashier was. \#terrible}
\hspace*{6mm} fear: 0.31\\[-8pt]


\noindent We believe that not vocalizing an outward judgment of the situation made the speaker appear more fearful.  
The HQT--NQT subset of our dataset will also be made separately, and freely, available as it may be of interest on its own, 
especially for the psychology and social sciences communities.





\begin{table*}[t]
\begin{center}
\resizebox{0.85\textwidth}{!}{
\begin{tabular}{lllll}
\hline
 & Twitter & Annotation &  Scope & Label \\ \hline
AFINN \cite{nielsen2011new} & Yes & Manual & Sentiment & Numeric \\ 
BingLiu \cite{Liu2004} & No & Manual & Sentiment & Nominal \\ 
MPQA \cite{Wilson05} & No & Manual & Sentiment & Nominal \\ 
NRC Affect Intensity Lexicon (NRC-Aff-Int) \cite{mohammad2017word} & Yes & Manual & Emotions & Numeric\\
NRC Word-Emotion Assn.\@ Lexicon (NRC-EmoLex) \cite{MohammadT13} & No & Manual & Emotions & Nominal\\ 
NRC10 Expanded (NRC10E) \cite{bravo2016determining} & Yes & Automatic & Emotions & Numeric \\ 
NRC Hashtag Emotion Association Lexicon (NRC-Hash-Emo) & Yes & Automatic & Emotions & Numeric \\ 
$\;\;\;\;\;$ \cite{COIN:COIN12024} &\\
NRC Hashtag Sentiment Lexicon (NRC-Hash-Sent)  \cite{MohammadSemEval2013} & Yes & Automatic & Sentiment & Numeric \\ 
Sentiment140 \cite{MohammadSemEval2013} & Yes & Automatic & Sentiment & Numeric \\ 
SentiWordNet \cite{Esuli06} & No & Automatic & Sentiment & Numeric \\ 
SentiStrength \cite{ThelwallBP12} & Yes & Manual & Sentiment & Numeric \\ \hline
\end{tabular}}
\end{center}
\vspace*{-2mm}
\caption{Affect lexicons used in our experiments.}
\label{tab:lex_prop}
\end{table*}

\section{Automatically Determining Tweet Emotion Intensity}
We now describe our 
regression system, which we use for 
obtaining benchmark prediction results on the new Tweet Emotion Intensity Dataset (Section \ref{sec:exp1}) and 
for determining the extent to which two emotions are correlated (Section \ref{sec:exp2}). 



 {\bf Regression System} We implemented a package called AffectiveTweets for the Weka machine learning workbench \cite{Wekapaper} that provides a collection of filters for extracting state-of-the-art 
 features from tweets for sentiment classification and other related tasks. These include features used in \citet{Kiritchenko2014} and \citet{MohammadSK17}.\footnote{\citet{Kiritchenko2014} describes the NRC-Canada system which ranked first in three sentiment shared tasks: SemEval-2013 Task 2, SemEval-2014 Task 9, and SemEval-2014 Task 4. \citet{MohammadSK17} describes a stance-detection system that outperformed submissions from all 19 teams that participated in SemEval-2016 Task 6.}
We use the package for calculating feature vectors from our emotion-intensity-labeled tweets and train Weka regression models on this transformed data.
We used an $L_{2}$-regularized $L_{2}$-loss SVM regression model with the  regularization parameter $C$ set to 1, implemented in LIBLINEAR\footnote{{http://www.csie.ntu.edu.tw/$\sim$cjlin/liblinear/}}. 
The features used:\footnote{See Appendix (A.3) for further implementation details.}


\noindent {\it a. Word N-grams (WN)}: presence or absence of word n-grams from $n=1$ to $n=4$. \\[-15pt]

\noindent {\it b. Character N-grams (CN)}: presence or absence of character n-grams from $n=3$ to $n=5$. 
\\[-15pt]

\noindent {\it c. Word Embeddings (WE)}: an average of the word embeddings of all the words in a tweet. We calculate individual word embeddings using the negative sampling skip-gram model implemented in \textit{Word2Vec} \cite{mikolov2013efficient}. 
Word vectors are trained from ten million English tweets taken from the Edinburgh Twitter Corpus \cite{Petrovic2010}. We set {\it Word2Vec} parameters: window size: 5; number of dimensions: 400.\footnote{\ed{Optimized for the task of word--emotion classification on an independent dataset \cite{bravo2016determining}.}}\\[-15pt]

\noindent {\it d. Affect Lexicons (L)}: we 
use the lexicons shown in Table \ref{tab:lex_prop}, by aggregating the information for all the words in a tweet. 
If the lexicon provides nominal association labels (e.g, positive, anger, etc.), then the number of words in the tweet matching each class are counted. 
If the lexicon provides numerical scores, the individual scores for each class are summed. 
These resources differ 
 according to: whether the lexicon includes Twitter-specific terms, whether the words were manually or automatically annotated, whether the words were annotated for sentiment or emotions,
and whether the affective associations provided are nominal or numeric. (See Table~\ref{tab:lex_prop}.)\\[-15pt] 

{\bf Evaluation} 
We calculate the Pearson correlation coefficient (r) between the scores produced by the automatic system on the test sets and the gold intensity scores 
to determine the extent to which the output of the 
system matches the results of human annotation.\footnote{We also determined Spearman rank correlations but these were inline with the results obtained using Pearson.}
Pearson coefficient, which  measures linear correlations between two variables, produces scores from -1 (perfectly inversely correlated) to 1 (perfectly correlated).
A score of 0 indicates no correlation.

\begin{table}[t!]
\resizebox{0.48\textwidth}{!}{
\begin{tabular}{lrrrrr}
\hline
 \multicolumn{2}{r}{\bf anger} &\bf fear &\bf joy & \bf sad. &\bf avg.\\ \hline
{\it Individual feature sets}\\
$\;\;\;$ word ngrams (WN) 						& 0.42 & 0.49 & 0.52 & 0.49 & 0.48 \\ 
$\;\;\;$ char.\@ ngrams (CN) 						& 0.50 & 0.48 & 0.45 & 0.49 & 0.48 \\ 
$\;\;\;$ word embeds.\@ (WE) 						& 0.48 & 0.54 & 0.57 & 0.60 & 0.55 \\ 
$\;\;\;$ all lexicons (L) 						&\bf 0.62 &\bf 0.60 &\bf 0.60 &\bf 0.68 &\bf 0.63 \\ 
$\;\;\;$ \it Individual Lexicons &\\
$\;\;\;\;\;\;$ AFINN 					& 0.48 & 0.27 & 0.40 & 0.28 & 0.36 \\ 
$\;\;\;\;\;\;$ BingLiu 				& 0.33 & 0.31 & 0.37 & 0.23 & 0.31 \\ 
$\;\;\;\;\;\;$ MPQA 					& 0.18 & 0.20 & 0.28 & 0.12 & 0.20 \\ 
$\;\;\;\;\;\;$ NRC-Aff-Int 					& 0.24 & 0.28 & 0.37 & 0.32 & 0.30 \\
$\;\;\;\;\;\;$ NRC-EmoLex 					& 0.18 & 0.26 & 0.36 & 0.23 & 0.26 \\ 
$\;\;\;\;\;\;$ NRC10E 					& 0.35 & 0.34 & 0.43 & 0.37 & 0.37 \\ 
$\;\;\;\;\;\;$ NRC-Hash-Emo 			&\bf 0.55 &\bf 0.55 &\bf 0.46 & 0.54 &\bf 0.53 \\ 
$\;\;\;\;\;\;$ NRC-Hash-Sent 			& 0.33 & 0.24 & 0.41 & 0.39 & 0.34 \\ 
$\;\;\;\;\;\;$ Sentiment140 			& 0.33 & 0.41 & 0.40 & 0.48 & 0.41 \\ 
$\;\;\;\;\;\;$ SentiWordNet 			& 0.14 & 0.19 & 0.26 & 0.16 & 0.19 \\ 
$\;\;\;\;\;\;\;$SentiStrength 			& 0.43 & 0.34 &\bf 0.46 &\bf 0.61 & 0.46 \\
{\it Combinations} &\\
$\;\;\;$ WN + CN + WE 			& 0.50 & 0.48 & 0.45 & 0.49 & 0.48 \\ 
$\;\;\;$ WN + CN + L 			& 0.61 & 0.61 & 0.61 & 0.63 & 0.61 \\ 
$\;\;\;$ WE + L 					& \textbf{0.64} & 0.63 & \textbf{0.65} & \textbf{0.71} & \textbf{0.66}\\
$\;\;\;\;$WN + WE + L 			& 0.63 & \textbf{0.65} & \textbf{0.65} & 0.65 & 0.65 \\ 
$\;\;\;$ CN + WE + L 			& 0.61 & 0.61 & 0.62  & 0.63 & 0.62 \\ 
$\;\;\;$ WN + CN + WE + L 		& 0.61 & 0.61 & 0.61 & 0.63 & 0.62 \\ 
\hline
\end{tabular}
}
\caption{Pearson correlations (r) of emotion intensity predictions with gold scores. Best results for each column are shown in bold: highest score by a 
feature set, highest score using a single lexicon, and highest score using feature set combinations.}
\label{tab:reg_res_full}
 \vspace*{-2mm}
\end{table}

\subsection{Supervised Regression and Ablation}
\label{sec:exp1}

We  developed our system by training on the official training sets and applying the learned models to the development sets. Once system parameters were frozen, the system trained on the combined training and development corpora. These models were applied to the official test sets.
Table  \ref{tab:reg_res_full} shows the results obtained on the test sets using
various features, individually and in combination. The last column `avg.' shows the
macro-average of the  correlations for all of the emotions.  

Using just character or just word n-grams leads to results around 0.48, suggesting that they are reasonably good indicators of emotion intensity by themselves.  (Guessing the  intensity scores at random between 0 and 1 is expected to get correlations close to 0.)
Word embeddings produce
statistically significant 
improvement over the ngrams (avg.\@ r = 0.55).\footnote{We used the
Wilcoxon signed-rank test at 0.05 significance level calculated from ten random partitions of the data,
for all the significance tests reported in this paper.}
Using features drawn from affect lexicons produces results ranging from  
avg.\@ r = 0.19  with SentiWordNet to  avg.\@ r =  0.53 with NRC-Hash-Emo.  
Combining all the lexicons  leads to 
statistically significant 
improvement over individual lexicons (avg.\@ r = 0.63). 
Combining the different kinds of features  leads to even higher scores, with the best overall result obtained using  
word embedding and lexicon features (avg.\@ r = 0.66).\footnote{The increase from 0.63 to 0.66 is statistically significant.}
The feature space formed by all the lexicons together 
is the strongest single feature category. The results also show that some features such as character ngrams are redundant in the presence of certain other features.

Among the lexicons, NRC-Hash-Emo is the most predictive single lexicon.  Lexicons that include Twitter-specific entries, 
lexicons that include intensity scores, and lexicons that label emotions and not just sentiment, tend to be more  predictive on this task--dataset combination. 
\ed{NRC-Aff-Int has real-valued fine-grained word--emotion association scores for all the words in NRC-EmoLex that were marked as being associated with anger, fear, joy, and sadness.\footnote{\ed{\scalebox{0.95}{http://saifmohammad.com/WebPages/AffectIntensity.htm}}}
Improvement in scores obtained  using NRC-Aff-Int  over the scores obtained using NRC-EmoLex also show that using fine intensity scores of word-emotion association are beneficial for tweet-level emotion intensity detection.}
The correlations for anger, fear, and joy 
 are similar 
(around 0.65), but the correlation for sadness is markedly higher (0.71).  We can observe from 
Table \ref{tab:reg_res_full}  
that this  boost in performance for sadness is to some extent due to word embeddings, but is more so due to  lexicon features, 
especially those from  SentiStrength. 
SentiStrength focuses solely on positive and negative classes, but provides numeric scores for each. 


\subsubsection{Moderate-to-High Intensity Prediction}
In some applications, 
it may be more important for a system to correctly determine emotion intensities in the higher range of the scale than in the lower range of the scale. 
To assess performance in the moderate-to-high range of the intensity scale, we calculated correlation scores over a subset of the test data formed by taking only those instances with gold emotion intensity scores $\geq 0.5$.

Table~\ref{tab:reg_res_range} shows the results.  Firstly, the correlation scores are in general lower here in the 0.5 to 1 range of intensity scores than in the experiments over the full intensity range.  This is simply because this is a harder task as now the systems do not benefit by making coarse distinctions over whether a tweet is  in the lower range or in the higher range.  
Nonetheless, we observe that  many of the broad patterns of results stay the same, with some differences.
Lexicons still play a crucial role, however, now embeddings and word ngrams are  not far behind. SentiStrength seems to be less useful in this range, suggesting that its main benefit was separating low- and high-intensity sadness words.  NRC-Hash-Emo is still the  source of the most predictive lexicon features.



\begin{table}[t!]
\resizebox{0.48\textwidth}{!}{
\begin{tabular}{lrrrrr}
\hline
 \multicolumn{2}{r}{\bf anger} &\bf fear &\bf joy & \bf sad. &\bf avg.\\ \hline
{\it Individual feature sets}\\
$\;\;\;$ word ngrams (WN) 		& 0.36 & 0.39 &\bf 0.38 & 0.40 & 0.38\\ 
$\;\;\;$ char.\@ ngrams (CN) 	& 0.39 & 0.36 & 0.34 & 0.34 & 0.36  \\ 
$\;\;\;$ word embeds.\@ (WE) 	& 0.41 & 0.42 & 0.37 & 0.51 & 0.43 \\ 
$\;\;\;$ all lexicons (L) 			&\bf 0.48 &\bf 0.47 & 0.29  &\bf 0.51  &\bf 0.44 \\ 
\multicolumn{6}{l}{$\;\;\;$ \it Individual Lexicons}\\
\multicolumn{6}{l}{$\;\;\;$ \it (some low-score rows not shown to save space)}\\
$\;\;\;\;\;\;$ AFINN 					& 0.31 & 0.06 & 0.11 & 0.05 & 0.13 \\ 
$\;\;\;\;\;\;$ BingLiu 					& 0.31 & 0.06 & 0.11 & 0.05 & 0.13 \\ 
$\;\;\;\;\;\;$ NRC10E 					& 0.27 & 0.14 &\bf 0.25 & 0.30 & 0.24  \\ 
$\;\;\;\;\;\;$ NRC-Hash-Emo 		&\bf 0.43 &\bf 0.39 & 0.15 &\bf 0.44 & \bf 0.35\\ 
$\;\;\;\;\;\;$ Sentiment140 			& 0.18 & 0.24 & 0.09 & 0.32 & 0.21\\ 
$\;\;\;\;\;\;$ SentiStrength 			& 0.23 & 0.04  & 0.19  & 0.34 & 0.20\\
{\it Combinations} &\\
 $\;\;\;$ WN + CN + WE 			& 0.37 & 0.35 & 0.33 & 0.34 & 0.35 \\ 
 $\;\;\;$ WN + CN + L 			& 0.44 & 0.45 & 0.34 & 0.43 & 0.41 \\ 
$\;\;\;$  WE + L 			& 0.51 & 0.49 & 0.38 & \textbf{0.54} &\textbf{0.48}\\
$\;\;\;$ WN + WE + L 			& \textbf{0.51} & \textbf{0.51} & \textbf{0.40} & 0.49 & 0.47 \\ 
 $\;\;\;$ CN + WE + L 			& 0.45 & 0.45 & 0.34  & 0.43 & 0.42 \\ 
$\;\;\;$ WN + CN + WE + L 		& 0.44 & 0.45 & 0.34 & 0.43 & 0.42 \\ 
 \hline
\end{tabular}}
\caption{Pearson correlations 
on a subset of the test set where gold scores $\geq 0.5$.}
\label{tab:reg_res_range}
\vspace*{-3mm}
\end{table}

\subsection{Similarity of Emotion Pairs}
\label{sec:exp2}
Humans are capable of hundreds of emotions, and some are closer to each other than others. 
One reason why certain emotion pairs may be perceived as being close is that their manifestation in language
is similar, for example, similar words and expression are used when expressing both emotions. 
We quantify this similarity of linguistic manifestation 
by using the Tweet Emotion Intensity dataset for the following experiment:
we train our 
regression system (with features WN + WE + L) on the training data for one emotion 
and evaluate predictions on the test data for a different emotion. 

Table~\ref{tab:reg_trans_full} shows the results. The numbers in the diagonal  are results obtained using
training and test data pertaining to the same emotion. These results are upperbound benchmarks for the
non-diagonal results, which are expected to be lower.  We observe that negative emotions are positively
correlated with each other and negatively correlated with the only positive emotion (joy).  The absolute
values of these correlations  go from $r=0.23$ to $r=0.65$. This shows that  all of the emotion pairs are
correlated at least to some extent, but that  in some cases, for example, when learning from fear data and
predicting sadness scores, one can obtain  results ($r=0.63$) close to the upperbound benchmark ($r=0.65$).\footnote{0.63 and 0.65 are not statistically significantly different.}
Note also that the correlations are asymmetric. This means that even though one emotion may be strongly
predictive of another, the predictive power need not be similar in the other direction.  
We also  found that training on a simple combination of both the fear and sadness data
and using the model to predict sadness obtained a correlation of 0.67 (exceeding the score obtained with
just the sadness training set).\footnote{0.67--0.63 difference is  statistically significantly different, 
but 0.67--0.65 and 0.65--0.63 differences are not.}
Domain adaptation may provide further gains. 

\begin{table}[t]
\begin{center}
{\small
\begin{tabular}{lrrrr}
\hline
 					&\multicolumn{4}{c}{\textbf{Test On}}\\
\textbf{Train On} & anger & fear & joy & sadness \\ \hline
anger & 0.63 & 0.37 & -0.37 & 0.45 \\ 
fear & 0.46 & 0.65 & -0.39 & 0.63 \\ 
joy & -0.41 & -0.23 & 0.65 & -0.41 \\ 
sadness & 0.39 & 0.47 & -0.32 & 0.65 \\ \hline
\end{tabular}}
\end{center}
\vspace*{-2mm}
\caption{Emotion intensity transfer Pearson correlation on all target tweets.}
\label{tab:reg_trans_full}
\vspace*{-2mm}
\end{table}

To summarize, the experiments in this section show
the extent to which two emotion are similar as per their manifestation in language. For the four emotions studied here, the similarities
vary from small (joy with fear) to considerable (fear with sadness). Also, the similarities are asymmetric. 
We also show that in some cases it is beneficial to use the training data for another emotion to supplement the training data for the emotion of interest.
A promising avenue of future work is 
to test theories of emotion composition: e.g,
whether optimism is indeed a combination of joy and anticipation, whether awe if fear and surprise, and so
on, as some have suggested \cite{Plutchik80}.

\section{Conclusions}
We created the first 
emotion intensity dataset for tweets.  We used best--worst scaling to
improve annotation consistency and obtained fine-grained scores.  We showed that emotion-word hashtags often
impact emotion intensity, often conveying a more intense emotion.  We created a benchmark regression system
and conducted experiments to show that affect lexicons, especially those with fine word--emotion association
scores, are 
useful in determining emotion intensity. Finally, we showed the extent to which
emotion pairs are correlated, and that the correlations are asymmetric---e.g., fear is strongly indicative of
sadness, but sadness is only moderately indicative of fear.

\section*{Acknowledgment}
\vspace*{-2mm}
We thank Svetlana Kiritchenko and Tara Small for helpful discussions.

\bibliography{maxdiff}
\bibliographystyle{acl_natbib}


\newpage

\appendix
\section{Appendix}

\subsection{Best--Worst Scaling Questionnaire used to Obtain Emotion Intensity Scores}
The BWS questionnaire used for obtaining fear annotations is shown below. 

{
\noindent\makebox[\linewidth]{\rule{0.5\textwidth}{0.4pt}}

\noindent {\bf Degree Of Fear In English Language Tweets}\\[-12pt]

\noindent The scale of fear can range from not fearful at all (zero amount of fear) to extremely fearful. One can often infer the degree of fear felt or expressed by a person from what they say. The goal of this task is to determine this degree of fear. Since it is hard to give a numerical score indicating the degree of fear, we will give you four different tweets and ask you to indicate to us:
\begin{itemize}
\item Which of the four speakers is likely to be the MOST fearful, and\\[-20pt] 
\item Which of the four speakers is likely to be the LEAST fearful. 
\end{itemize}


\noindent {\bf Important Notes}
\begin{itemize}
\item This task is about fear levels of the speaker (and not about the fear of someone else mentioned or spoken to).\\[-20pt]
\item If the answer could be either one of two or more speakers (i.e., they are likely to be equally fearful), then select any one of them as the answer.\\[-20pt]
\item Most importantly, try not to over-think the answer. Let your instinct guide you.\\[-13pt]
\end{itemize}

\noindent {\bf EXAMPLE}\\[-8pt]

\noindent Speaker 1: {\it Don't post my picture on FB \#grrr}\\
Speaker 2: {\it If the teachers are this incompetent, I am afraid what the results will be.}\\
Speaker 3: {\it Results of medical test today \#terrified}\\
Speaker 4: {\it Having to speak in front of so many people is making me nervous.}\\

\noindent Q1. Which of the four speakers is likely to be the MOST fearful?\\ 
-- Multiple choice options: Speaker 1, 2, 3, 4 --\\
\noindent Ans: Speaker 3\\[-8pt]

\noindent Q2. Which of the four speakers is likely to be the LEAST fearful?\\ 
-- Multiple choice options: Speaker 1, 2, 3, 4 --\\
\noindent Ans: Speaker 1\\[-20pt]

\noindent\makebox[\linewidth]{\rule{0.5\textwidth}{0.4pt}}

}
\noindent The questionnaires for other emotions are similar in structure.
In a post-annotation survey, the respondents gave the task high scores for clarity of instruction (4.2/5) despite noting that the task itself requires some non-trivial amount of thought (3.5 out of 5 on ease of task).

\subsection{Use of Emotion Word Hashtags}
Emotion word hashtags (e.g., {\it \#angry, \#fear}) have been used 
 to search and compile sets of tweets that are likely to convey the emotions of interest. 
Often, these tweets are used in one of two ways:
1. As noisy training data for distant supervision  \cite{pak2010twitter,Mohammad12,suttles2013distant}.
2. As data that is manually annotated for emotions to create training and test datasets suitable for machine learning \cite{roberts2012empatweet,qadir2014learning,Mohammad2015elec}.\footnote{Often,
the query term is removed from the tweet so as to erase obvious cues for a classification task.}
We use emotion word hashtag to create annotated data similar to `2', however, we use them to create separate emotion intensity datasets for each emotion.
We also 
examine the impact of emotion word hashtags on emotion intensity. 
This has not been studied before, even though there is work on
learning  hashtags associated with particular emotions \cite{qadir2013bootstrapped}, 
and on showing that some emotion word hashtags are strongly indicative of the presence of an emotion in the rest of the tweet, 
whereas others are not \cite{kunneman2014predictability}.

\begin{figure*}[t]
\centering
 \includegraphics[width=5.6in]{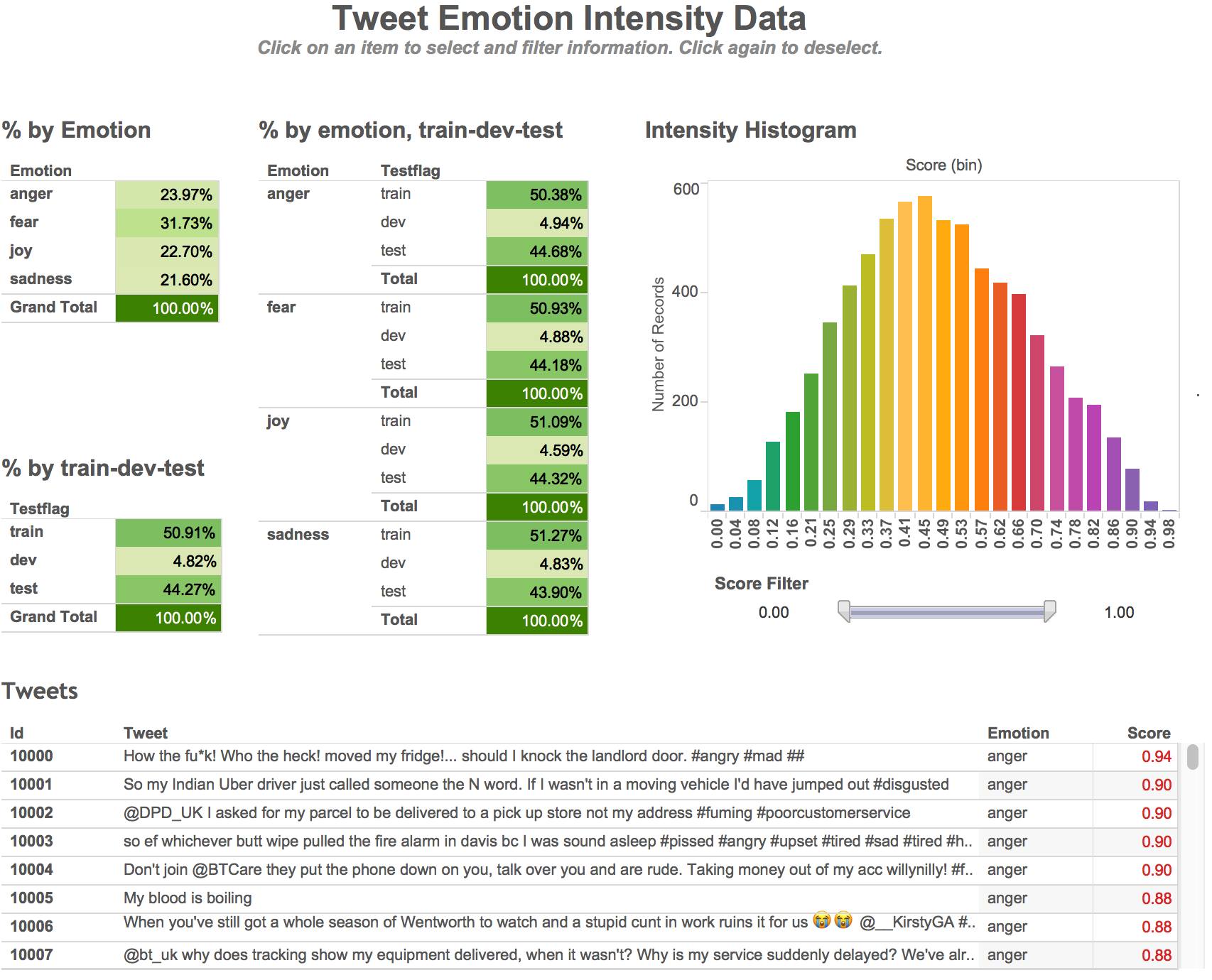}
\caption{Screenshot of the interactive visualization to explore the Tweet Emotion Intensity Dataset.\\
\ed{Available at: http://saifmohammad.com/WebPages/EmotionIntensity-SharedTask.html}}
\label{fig:viz1}
\end{figure*}

\subsection{AffectiveTweets Weka Package}

AffectiveTweets includes five filters for converting tweets into feature vectors that can be fed into the large collection of machine learning algorithms implemented within Weka. The package is installed using the \textit{WekaPackageManager} and can be used from the Weka GUI or the command line interface. It uses the \textit{TweetNLP} library \citep{Gimpel11} for tokenization and POS tagging. The filters are described as follows.\\[-23pt]
\begin{itemize}
\item \textit{TweetToSparseFeatureVector} filter: calculates the following sparse features: word n-grams (adding a NEG prefix to words occurring in negated contexts), 
character n-grams (CN),  POS tags, and  Brown word clusters.\footnote{The scope of negation was determined by a simple heuristic: from the occurrence of a negator word up until a punctuation mark or end of sentence. We used a list of 28 negator words such as {\it no, not, won't} and {\it never}.}\\[-24pt]
\item \textit{TweetToLexiconFeatureVector} filter:  calculates features from a fixed list of affective lexicons.\\[-20pt]
\item \textit{TweetToInputLexiconFeatureVector}: calculates features from any lexicon. The input lexicon can have multiple numeric or nominal word--affect associations.\\[-20pt]  
\item \textit{TweetToSentiStrengthFeatureVector} filter: calculates positive and negative sentiment intensities for a tweet using the SentiStrength lexicon-based method \cite{ThelwallBP12}\\[-20pt]
\item \textit{TweetToEmbeddingsFeatureVector} filter: calculates a tweet-level feature representation using pre-trained word embeddings supporting the following aggregation schemes:  average of word embeddings;  addition of word embeddings; and concatenation of the first $k$ word embeddings in the tweet. The package also provides \textit{Word2Vec's} pre-trained word embeddings. 
\end{itemize}
\noindent Additional filters for creating affective lexicons from tweets and support for distant supervision are currently under development.

\subsection{An Interactive Visualization to Explore the Tweet Emotion Intensity Dataset}
We created an interactive visualization to allow ease of exploration of this new dataset.  
The visualization has several components:\\[-20pt]
\begin{enumerate}
\item Tables showing the percentage of instances in each of
the emotion partitions (train, dev, test). Hovering over a row shows the corresponding number of instances.
Clicking on an emotion filters out data from all other emotions, in all visualization components. Similarly, one can click on just the train, dev, or test
partitions to view information just for that data. Clicking again deselects the item.\\[-22pt]
\item A histogram of emotion intensity scores. A slider that one can use to view only those tweets within a certain score range.\\[-22pt]
\item The list of tweets, emotion label, and emotion intensity scores. \\[-20pt]
\end{enumerate}
\noindent 
One can use filters in combination. For e.g., clicking on fear, test data, and setting the slider for the 0.5 to 1 range,
shows information for only those fear--testdata instances with scores $\ge$ 0.5.

\end{document}